\documentclass{AAIML}

\usepackage[table]{xcolor}
\usepackage{amsfonts}
\usepackage{amsmath}
\usepackage{amssymb}
\usepackage[linesnumbered,ruled,vlined]{algorithm2e}
\SetKwComment{Comment}{$\triangleright$\ }{}

\usepackage{booktabs} % For better table rules (\toprule, \midrule, \bottomrule)
\usepackage{tabularx} % For tables with specified width
\usepackage{array}    % For column specifications
\usepackage{url}      % For handling URLs if needed

\usepackage{float}
\usepackage{listings}

\aaimlheading{1}{1}{41-46}{4/00}{10/00}{00-000}{Jeongseok Kim and Kangjin Kim}

\ShortHeadings{Intent-Driven UAM Rescheduling}{Jeongseok Kim and Kangjin Kim}
\firstpageno{41}

\mycitation{J. Kim and K. Kim}{Intent-Driven UAM Rescheduling.}{Advances in Artificial Intelligence and Machine Learnin.}{2025;8(6):6.}{}

\newcommand{\figtoybaseline}{Figure~\ref{fig:toy:baseline}}
\newcommand{\figtoyrescheduled}{Figure~\ref{fig:toy:rescheduled}}

\newcommand{\vect}[1]{\mathbf{#1}}

\newtheorem{problem}{Problem}

\usepackage{paralist}

\usepackage{etoolbox} % For \patchcmd
\usepackage{subcaption}

\ifundef{\mho}{\newcommand{\mho}{\textit{unknown}}}{} % Example definition
\ifundef{\rhd}{\newcommand{\rhd}{\triangleright}}{} % Example definition
\newcommand{\textproc}[1]{\textsc{#1}} % Formatting for procedure names
\newcommand{\textfunc}[1]{\texttt{\footnotesize #1}} % Formatting for function names

\begin{document}

\title{Intent-Driven UAM Rescheduling}

\author{\name Jeongseok Kim \email justin@cleverplant.com \\
       \addr Cleverplant \\
       Seoul, 04595, Republic of Korea \\
       \AND
       \name Kangjin Kim \email kangjinkim@cdu.ac.kr \\
       \addr  Department of Drone Systems \\
       Chodang University \\
       Jeollanam-do, 58530, Republic of Korea}
       
\editor{Kangjin Kim}

\maketitle

\begin{abstract}
Due to the restricted resources, efficient scheduling in vertiports has received much more attention in the field of Urban Air Mobility (UAM). For the scheduling problem, we utilize a Mixed Integer Linear Programming (MILP), which is often formulated in a resource-restricted project scheduling problem (RCPSP). In this paper, we show our approach to handle both dynamic operation requirements and vague rescheduling requests from humans. Particularly, we utilize a three-valued logic for interpreting ambiguous user intents and a decision tree, proposing a newly integrated system that combines Answer Set Programming (ASP) and MILP. This integrated framework optimizes schedules and supports human inputs transparently. With this system, we provide a robust structure for explainable, adaptive UAM scheduling.
\end{abstract}


\section{Introduction}

Urban Air Mobility (UAM) introduces a new traffic paradigm that requires efficient management of vertical take-off and landing (VTOL) resources. In this context, scheduling is formulated in a resource-constrained project scheduling problem (RCPSP), which is often solved through a mixed integer linear programming (MILP) approach. The dynamic nature of UAM, however, also requires human operators to fluidly control the schedules in order to meet the real-time requirements.

Problems mostly can occur when  the operator's request is ambiguous. In such a case, automated systems have to interpret the underlying intent more precisely before revising the schedule. This is because misunderstanding the intent is able to affect the efficiency and the trust negatively. To address this, we propose a system that combines a MILP-based scheduler with an Answer Set Programming (ASP) module for intent interpretation. This module uniquely uses three-valued logic and decision trees to resolve ambiguity.

In order to make it easy to understand the reasoning process of the system under the schedule revision, we utilize a declarative approach of ASP. This can cover both explicit and vague inputs and reason on its proper handling, keeping the interpretability. This improves the system's reactivity, building user trust and meeting the explainable HRI rules.

We previously studied human interaction in Urban Air Traffic Management (UATM) systems \cite{Kim2023Agent3C, Kim2023AAIML, Woo2023, Kim2024ICICT}. These studies mainly focused on handling route changes and reactive event processing (such as detours and vertiport closures) using knowledge representation and reasoning. Extending this direction, we recently studied the vertiport resource scheduling problem. Particularly, we utilized ASP in order to meet the constraints of various stakeholders within RCPSP formulations \cite{Kim2025}. This paper, based on the study, focuses on vertiport rescheduling, handling the issues of interpreting ambiguous requests from human operators in dynamic operational settings.

The contributions of this paper are threefold: firstly, we formulate UAM scheduling as an RCPSP for the purpose of utilizing the MILP solver to compute the schedule, secondly, we introduce an ASP-based intent recognition mechanism that utilizes three-valued logic, and lastly, we propose an explainable system that integrates human intent and automated scheduling. In the following sections, we cover related works, approaches, experiments, and conclusions in detail.

\section{Related Works}

Vertiports play the role of hubs that are built for operating eVTOLs in urban settings. Research in \cite{Safwat2024, Wang2021, Li2023} shows that the efficacy of UAM largely depends on the proper scheduling. Essentially, authors in \cite{Wei2021, Thipphavong2018} provided that scheduling these jobs is complex due to limited resources, integration with urban systems, and safety constraints . This problem, therefore, is formulated as an RCPSP. This problem is in the NP-hard class, which is to optimize the makespan while keeping the resource and precedence constraints (refer to \cite{Zaman2018, Hosseinian2019}). In such constraints, MILP has been broadly utilized \cite{Gholamnejad2020, Liu2021} even though it requires a high computational cost in order to derive optimal schedules in various domains \cite{Mouret2009}. In addition, meta-heuristics and various optimization strategies have been developed in order to handle uncertainties \cite{Duan2010, Mahmud2021, Alipouri2021, Davari2018, Nemmich2019}.

The research in \cite{Li2017, Hancock2013, Han2013} indicates that it becomes more challenging to accurately recognize the users' intent in coordination when integrating human operators into an automated scheduling system. Particularly, resolving the ambiguity is the key issue when human inputs are unclear or not properly defined. Research in \cite{Le2022, Liu2016, Berawi2020} proves that methodologies utilizing multi-valued logic systems and decision trees show some potential to manage the uncertainties in decision-making contexts.

The declarative programming paradigm enables the resolution of complex problems within a systematic and verifiable framework. The effectiveness of ASP, in particular, has been demonstrated in the domain of knowledge representation and reasoning. \cite{Kordjamshidi2022, Dodaro2022, Mazzotta2022, Calimeri2022} showed its expressiveness, enabling us to model complex problems such as scheduling and decision-making under uncertainty. This transparency aligns with the objectives of explainable AI (XAI) pursuits in the human-robot interaction domain. Providing a clear explanation for the system's decision-making process can build user trust, guarantee accountability, and lead to effective human-AI collaboration (refer to \cite{Wang2023, Kumar2024, Karran2022, Vemuri2023, Akula2022}). This research implements an explainable UAM scheduling system that is responsive to human intent, following ambiguity resolution rules through ASP.

\section{Preliminaries}
\subsection{RCPSP}

The Resource Constrained Project Schedule Problem (RCPSP) aims to find a schedule for tasks $J=\{0..n+1\}$ that minimizes the project completion time (deadline). This schedule is constrained by a priority constraint $S^0$ between tasks and a set of available renewable resources $R=\{1..m\}$. The limit $C_r$ of resources $R=\{1..m\}$ must be observed. Each job $j$ has a duration $p_j$ and requires a resource $r$ of unit $u^0_{jr}$ while the job is active. Jobs 0 and $n+1$ are dummy start and end jobs.

The general MILP formulation uses binary variables $x_{jt}$ where $x_{jt}=1$ if task $j$ starts at time $t$ and $0$ otherwise. The objective is to minimize the start time of the final task:

\begin{subequations}
	\label{eq:milp_model_condensed_reduced} % Changed label slightly to avoid duplicate if compiling both versions
	\begin{align}
		\min \; & \sum_{t=0}^{H} t \cdot x_{n+1, t} \\
		\text{s.t.} \; & \sum_{t=0}^{H-p_j} x_{jt} = 1 && \forall j \in J \label{eq:milp_c1_reduced} \\
		& \sum_{j \in J}\sum_{\tau = \max(0, t - p_j + 1)}^{t} u^0_{jr}\! \cdot x_{j\tau}\! \le\! C_r && \forall r, t \label{eq:milp_c2_reduced} \\
		& \sum_{t} t x_{st} - \sum_{t} t x_{jt} \ge p_j && \forall (j,\! s) \in S^0 \label{eq:milp_c3_reduced} \\
		& x_{jt} \in \{0, 1\} && \forall j, t \label{eq:milp_c4_reduced}
	\end{align}
\end{subequations}

\begin{figure}[H] % Make figure placement more flexible
					\centering
					\begin{subfigure}[b]{0.8\linewidth}
						\includegraphics[width=\linewidth]{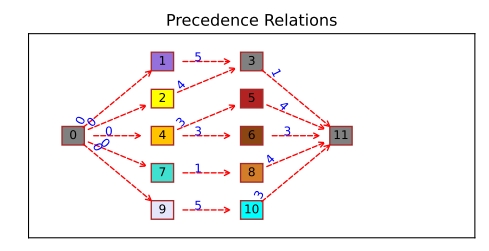}
						\caption{\spaceskip.15em Precedence Relations (Tasks 0 and 11 are dummies).}\label{fig:toy:trees}
						\vspace{0.8\baselineskip}
					\end{subfigure}
					
					\begin{subfigure}[b]{.493\linewidth}
						\includegraphics[width=\linewidth]{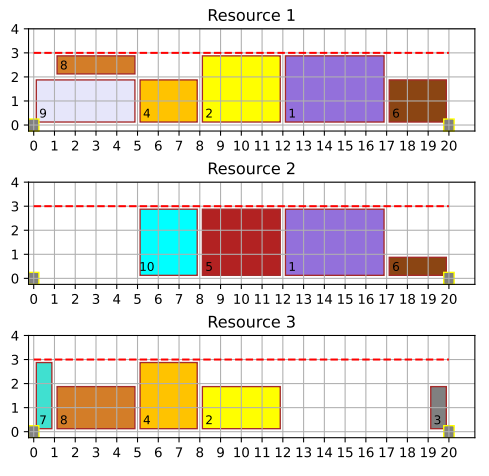}
						\caption{Resource Assignment: Baseline Solution.}\label{fig:toy:baseline}
					\end{subfigure}
					\begin{subfigure}[b]{.493\linewidth}
						\includegraphics[width=\linewidth]{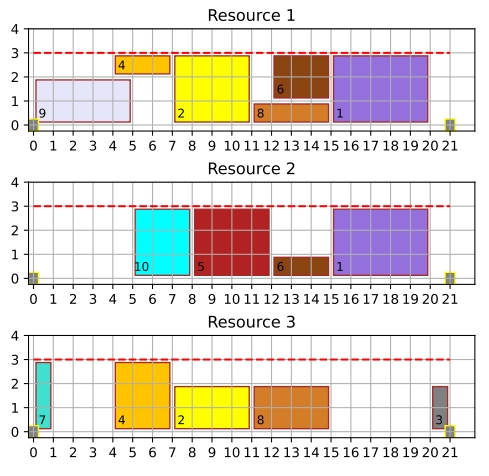}
						\caption{Resource Assignment: Rescheduled Solution.}\label{fig:toy:rescheduled}
					\end{subfigure}
					\vspace{0.4\baselineskip}
					\caption{Comparison of Baseline and Rescheduled Schedules for a Resource Change Request on Task 4.}
					\label{fig:toy_example_result}
				\end{figure}

Constraints ensure each task starts exactly once (\ref{eq:milp_c1_reduced}), resource capacities are not exceeded (\ref{eq:milp_c2_reduced}), and precedence relations are met (\ref{eq:milp_c3_reduced}). An example RCPSP instance is provided in Table~\ref{tab:rcpsp_example_condensed_reduced}. Figure \ref{fig:toy:baseline} is a visual representation of the schedule that corresponds to the related schedule.

% Use \small or \footnotesize if needed and adjust column widths
\begin{table}[h!]
	\centering
	\caption{Example RCPSP instance data (Referenced in \figtoybaseline).} % Adjusted figure ref
	\label{tab:rcpsp_example_condensed_reduced}
	%\footnotesize % Reduce font size slightly if needed
	\begin{tabularx}{0.7\columnwidth}{@{} c c c c c c @{}} % Use \columnwidth for width
		\toprule
		Task & Duration & \multicolumn{3}{c}{Required ($u_{jr}$)} & Successors \\
		\cmidrule(lr){3-5} % rule under resource columns only
		& ($p_j$) & R1 & R2 & R3 & \\
		\midrule
		1  & 5 & 3 & 3 &   & 3 \\
		2  & 4 & 3 &   & 2 & 3 \\
		3  & 1 &   &   & 2 &   \\
		4  & 3 & 2 &   & 3 & 5, 6 \\
		5  & 4 &   & 3 &   &   \\
		6  & 3 & 2 & 1 &   &   \\
		7  & 1 &   &   & 3 & 8 \\
		8  & 4 & 1 &   & 2 &   \\
		9  & 5 & 2 &   &   & 10 \\
		10 & 3 &   & 3 &   &   \\
		\bottomrule
	\end{tabularx}
\end{table}

\subsection{Rescheduling Options Overview}

A categorization of typical rescheduling strategies is provided in Table~\ref{tab:rescheduling_options_reduced}, which is based on the literature reviewed by \cite{Kuster2006}.
Our framework is centered on the following options that are highlighted: Partial Rescheduling and Resource Reallocation, more precisely the application of alterations to the start time of a single job or the quantity of resources that are required.

\begin{table}[htbp]
	\centering
	\footnotesize
	\caption{Rescheduling Options (Highlighted rows are covered in this paper)}
	\label{tab:rescheduling_options_reduced} % Changed label
	\begin{tabular}{|p{2.5cm}|p{2.5cm}|p{1.3cm}|p{1.3cm}|p{1.4cm}|}
		\hline
		\textbf{Option Category} & \textbf{Change Focus} & \textbf{Precedence Relations} & \textbf{Resource Assignment} & \textbf{Typical Impact} \\ \hline
		Full Rescheduling      & All future tasks       & Re-optimize entire graph   & Reassign all resources   & High cost; disrupts baseline \\ \hline
		\rowcolor{yellow!20} Partial Rescheduling   & Affected tasks only     & Modify only affected links  & Limited reallocation     & Balances stability and optimization \\ \hline
		Temporal Shift         & Start time adjustments  & Maintain original order    & Typically unchanged      & Minimal change; high stability \\ \hline
		Task Reordering        & Adjust sequence order   & Change order while preserving tasks & May indirectly affect resources & Improves performance; less stable \\ \hline
		\rowcolor{yellow!20} Resource Reallocation  & Modify resource amounts  & Usually preserves order    & Change resource assignments & Targets resource conflicts \\ \hline
		Structural Modification& Insert/remove tasks, change modes & Remove or add links & May involve reallocation   & Significant structural change \\ \hline
	\end{tabular}
\end{table}

	\subsection{UAM Context: Intent-Driven, Explainable Rescheduling}
%UAM operations necessitate frequent, rapid schedule adjustments. Our system integrates MILP-based RCPSP scheduling with an explainable Answer Set Programming (ASP) layer to manage these adjustments. Explainability is vital in human-robot interaction within UAM, fostering trust by providing clear justifications for system decisions \cite{Wang2023, Kumar2024}.
%
%Our approach is "intent-driven": it interprets the user's underlying goal when requesting a schedule change. By analyzing the request type (start time or resource adjustment) and potentially ambiguous user input, the ASP module determines the appropriate rescheduling action and explains its reasoning.
%
%\textbf{Scope:} This work focuses on single-task rescheduling within UAM operations via Partial Rescheduling or Resource Reallocation. This involves modifying a target task's start time or resource requirement. A key aspect is deciding whether to maintain the task's original precedence constraints or remove them to allow more isolated optimization during the adjustment.
%
%\textbf{Exclusions:} The scope does not cover global schedule reoptimization, reordering multiple tasks, or major structural changes involving adding/removing multiple tasks.
%-----
UAMs require frequent and quick rescheduling to maintain operations. Our system utilizes an explainable answer set programming (ASP), integrating MILP-based RCPSP scheduling to effectively handle these changes. Here explainability plays an important role in human-robot interactions within UAM because it helps build trust by providing transparent reasoning for system decisions. \cite{Wang2023, Kumar2024}.

Our approach is ``intent-driven": it interprets the user's underlying goal when requesting a schedule change. By analyzing the request type (start time or resource adjustment) and potentially ambiguous user input, the ASP module determines the appropriate rescheduling action and explains its reasoning.

\textbf{Scope:} The single task rescheduling within the UAM operation is the main focus of this study and is performed through partial rescheduling or resource reallocation. In this scope, we consider two requirement options: making adjustments to the start time of a specific job and the resources that are required. An important aspect of this coverage is to decide whether the task's original precedence constraints should be kept or if they should be removed in order to allow for independent optimization throughout the adjustment.

\textbf{Exclusions:} The scope does not cover global schedule reoptimization, reordering multiple tasks, or major structural changes involving adding/removing multiple tasks.

		\section{Interactive Intention Extraction Process}
			This section specifies in detail the process of extracting user intent from potentially ambiguous rescheduling requests and formulates the corresponding rescheduling problem.
		
%		\subsection{Ambiguous Rescheduling Request}
%		An initial (ambiguous) rescheduling request for a task $t^*$ is given as
%		\[Q =\ \bigl(\delta, \rho, \eta, \theta, \tau^*, R^{*}\bigr),
%		\]
%		where:
%		\begin{itemize}
%			\item $\delta \in \{unknown\} \cup \{0, 1\}$ indicates whether the new schedule should respect other tasks' start time in the current schedule (with 0 meaning "respect").
%			\item $\rho \in \{unknown\} \cup \{0, 1\}$ indicates whether the task should be independent of its precedence constraints (with 0 meaning "independent").
%			\item $\eta \in \{unknown\} \cup \{0, 1\}$ indicates whether the new schedule for a resource amount change request should respect other tasks' start time in the current schedule (with 0 meaning "respect").
%			\item $\theta \in \{unknown\} \cup \{0, 1\}$ indicates whether the task's precedence constraints are relaxed for the resource adjustment (with 0 meaning "relax").
%			
%			\item $\tau ^*\in \mathbb{Z}_{\ge 0}$ represents the desired start time for a task $t^*$.
%			\item $R^{*} \in \mathbb{Z}^{m}_{\ge 0}$ is the desired resource vector for a task $t^*$.
%		\end{itemize}
		
		\subsection{Ambiguous Rescheduling Request Representation}
%		An initial rescheduling request $Q$ for a specific task $t^*$ may contain ambiguities. We represent this request as a tuple:
		We consider that an initial rescheduling request $Q$ for a specific task $t^*$ may contain ambiguities. This request is represented as follows:
		\[ Q = (\delta, \rho, \eta, \theta, \tau^*, R^{*}) \]
		where:
		\begin{itemize}
\item $\delta \in \{\text{unknown}, 0, 1\}$ is to express the type of optimization for a time change request. $0$ is to respect start times of other tasks, only allowing local change; $1$ is to optimize globally; and 'unknown' represents ambiguity.
\item $\rho \in \{\text{unknown}, 0, 1\}$ is to express the type of relationship to be maintained for a time change request. $0$ is to relax precedence constraints for $t^*$; $1$ is to maintain the precedence relation; and 'unknown' represents ambiguity.
\item $\eta \in \{\text{unknown}, 0, 1\}$ is to express the type of optimization for a resource amount change request. $0$ is to respect start times of other tasks, only allowing local change; $1$ is to optimize globally; and 'unknown' represents ambiguity.
\item $\theta \in \{\text{unknown}, 0, 1\}$ is to express the type of relationship to be maintained for a resource amount change request. $0$ is to relax precedence constraints for $t^*$; $1$ is to maintain the precedence relation; and 'unknown' represents ambiguity.
\item $\tau^* \in \mathbb{Z}_{\ge 0}$ is the desired start time for task $t^*$.
\item $R^{*} \in \mathbb{Z}^{m}_{\ge 0}$ is the desired resource requirement vector for task $t^*$, where $m$ is the number of resource types.
\end{itemize}

If one of $\delta, \rho, \eta, \theta$ has 'unknown' as its value, an interactive process is required in order to clarify the user's true intent.

%		\subsection{Interactive Function Mapping}
%		Let the interaction history be modeled as
%		\[\mathcal{H} = \{\bigl(q^{(0)}, a^{(0)}\bigr), \bigl(q^{(1)}, a^{(1)}\bigr), \dots, \bigl(q^{(T)}, a^{(T)}\bigr)\}.
%		\]
%		Define the interactive mapping function
%		\[\mathcal{I}:Q \times \mathcal{H} \rightarrow \{0, 1\}^{4},
%		\]
%		which outputs a fully specified Boolean vector:
%		\[\mathbf{b}^{*} = \mathcal{I}\bigl(Q, \mathcal{H}\bigr) = (b^{(s)}_1, b^{(s)}_2, b^{(r)}_1, b^{(r)}_2).\]
%		
%		Here:
%		\begin{itemize}
%			\item $b^{(s)}_1$: equals 1 if an optimal reschedule is requested, by investigating and changing all other task's start time; 0 otherwise.
%			\item $b^{(s)}_2$: equals 1 if the task’s precedence constraints are to be kept for the start time decision; 0 if they are maintained.
%			\item $b^{(r)}_1$: equals 1 if an optimal reschedule is requested, by investigating and changing all other task's start time; 0 otherwise.
%			\item $b^{(r)}_2$: equals 1 if the task’s precedence constraints are to be kept for the resource adjustment; 0 if they are maintained.
%		\end{itemize}
%		
%		The interactive process proceeds iteratively:
%		\[\mathbf{b}^{(t+1)} = U\bigl(\mathbf{b}^{(t)}, q^{(t)}, a^{(t)}\bigr), t = 0, 1, \dots, \mathcal{T} - 1,
%		\]
%		starting from an initial guess $\mathbf{b}^{(0)}$ (where all components are initially undefined), until convergence:
%		\[\mathbf{b}^{*} = \lim_{t \rightarrow \mathcal{T}}\mathbf{b}^{(t)}.
%		\]
%		This final vector $\mathbf{b}^{*}$ captures the clarified user intentions regarding both start time and resource changes.

		\subsection{Interactive Intention Extraction}
        We resolve the ambiguity the initial request $Q$ has through interactions with the user. We utilize the concept of \textit{interaction history} in order to implement this interaction process. We denote this history as $\mathcal{H}$. $\mathcal{H}$ consists of a sequence of tuples, and each tuple captures an interaction.
        \begin{equation*}
		\mathcal{H} = [(q^{(0)}, a^{(0)}), (q^{(1)}, a^{(1)}), \dots, (q^{(T)}, a^{(T)})]
        \end{equation*}

Here, $q^{(t)}$ indicates the inquiry the system asks the user to clarify the ambiguity at step $t$, and $a^{(t)}$ is the corresponding response from the user. Every interaction step is appended as a new $(q, a)$ tuple to $\mathcal{H}$.

The system uses the interaction mapping function $\mathcal{I}$ to determine the user's fully clarified intent by utilizing the initial request $Q$ and the accumulated history $\mathcal{H}$:
        \begin{equation*}
		\mathcal{I}: Q \times \mathcal{H} \rightarrow \{0, 1\}^{4}
        \end{equation*}

The output is an intent vector, which is fully clarified as follows:

        \begin{equation*}
		\vect{b}^{*} = \mathcal{I}(Q, \mathcal{H}) = (b^{(s)}_1, b^{(s)}_2, b^{(r)}_1, b^{(r)}_2)
        \end{equation*}

		where:

		\begin{itemize}

\item $b^{(s)}_1$ (clarified $\delta$) is a boolean value to indicate the optimization type for a time change request. $1$ means that global optimization is intended, and $0$ means that local optimization is intended, respecting other tasks.

\item $b^{(s)}_2$ (clarified $\rho$) is a boolean value to indicate the precedence constraints for a time change request. $1$ means that the relation should be maintained; $0$ means that the relation is relaxed.

\item $b^{(r)}_1$ (clarified $\eta$) is a boolean value to indicate the optimization type for a resource amount change request. $1$ means that global optimization is intended, and $0$ means that local optimization is intended, respecting other tasks.

\item $b^{(r)}_2$ (clarified $\theta$) is a boolean value to indicate the precedence constraints for a resource amount change request. $1$ means that the relation should be maintained; $0$ means that the relation is relaxed.

		\end{itemize}

This vector $\vect{b}^{*}$ clearly defines conditions for the following rescheduling process.

		\section{Problem Formulation}

        We define a rescheduling process as applying a transition operator $\mathcal{R}$ to the baseline schedule $S_0$, given a request $Q$ for a task $t^*$.
	\begin{equation*}
		\mathcal{R}: (S_0, t^*, Q) \mapsto (S, E)	
	\end{equation*}	
	where:	
	\begin{itemize}
\item \textbf{Input}:
		\begin{enumerate}
			\item $S_0$ is the current baseline schedule.
			\item $t^* \in J$ is the target rescheduling task.
			\item $Q$ is a tuple that contains a potentially ambiguous rescheduling request.
		\end{enumerate}
\item \textbf{Process}:
		\begin{enumerate}
			\item \textbf{Intention Extraction}: It determines the clarified intent vector $\vect{b}^{*} = (b^{(s)}_1, b^{(s)}_2, b^{(r)}_1, b^{(r)}_2)$ through $\mathcal{I}(Q, \mathcal{H})$.
			\item \textbf{Parameter Update}: It updates RCPSP parameters for $t^*$. Firstly, it sets the desired start time for $t^*$ according to $\tau^*$. Secondly, it updates the resource requirement $u_{t^*r}$ according to $R^*$.
			\item \textbf{Constraint Adjustment}: It revises RCPSP constraints according to $\vect{b}^{*}$:
				\begin{itemize}
					\item \textit{Precedence}: It relaxes constraints for $t^*$ if $b^{(s)}_2 = 0$ (for the time change request). Otherwise, it keeps constraints for $t^*$.The same approach applies to resource amount change requests.
					\item \textit{Scope}: It keeps start times for unrelated tasks if $b^{(s)}_1 = 0$ (for the time change request). Otherwise, it globally optimizes start times. The same approach applies to resource amount change requests.\end{itemize}
			\item \textbf{Rescheduling Optimization}: It solves the modified RCPSP instance for the new schedule $S$.
			\item \textbf{Explanation Generation}: It produces $E$ to justify $S$ according to $Q$, $\vect{b}^{*}$, and the result.
		\end{enumerate}
\item \textbf{Output}:
		\begin{itemize}
			\item $S$: The new, updated schedule.
			\item $E$: An explanation of the rescheduling action.
		\end{itemize}
\end{itemize}

%		\paragraph{Remark on Notation.} The intent variables $b^{(s)}_1, b^{(s)}_2, b^{(r)}_1, b^{(r)}_2$ specify scope (1) and precedence (2) handling for start-time (s) and resource (r) requests. For conciseness, when the request type is clear (start-time or resource), we will simplify this to $b_1$ (scope) and $b_2$ (precedence), dropping the $(s)$ or $(r)$ superscript. The full notation $\vect{b}^{*} = (b^{(s)}_1, b^{(s)}_2, b^{(r)}_1, b^{(r)}_2)$ is used when necessary for clarity or referring to the complete intent vector.
		\paragraph{Remark on Notation.} The intent variables $b^{(s)}_1, b^{(s)}_2, b^{(r)}_1, b^{(r)}_2$ specify scope (1) and precedence (2) handling for start-time (s) and resource (r) requests. For conciseness, when the request type is clear (start-time or resource), we will simplify this to $b_1$ (scope) and $b_2$ (precedence).

		\subsection{Problem Statement}
%		\begin{problem}
%			Given a baseline schedule $S_0$ and a reschedule request $Q$ for a task $t^*$, compute a new schedule $S$, and explanation $E$ by interpreting the intention of $Q$ and doing the rescheduling process.
%		\end{problem}

		\begin{problem}
			Given a baseline schedule $S_0$ and a (potentially ambiguous) reschedule request $Q$ for a task $t^*$, compute a new schedule $S$ and a corresponding explanation $E$ by first interactively clarifying the user's intent $\vect{b}^{*}$ from $Q$ via $\mathcal{H}$, and then executing the intent-driven rescheduling process $\mathcal{R}$.
		\end{problem}

		\subsection{Solution Overview}
\begin{figure}[H]
			\centering
			\includegraphics[width=0.7\linewidth]{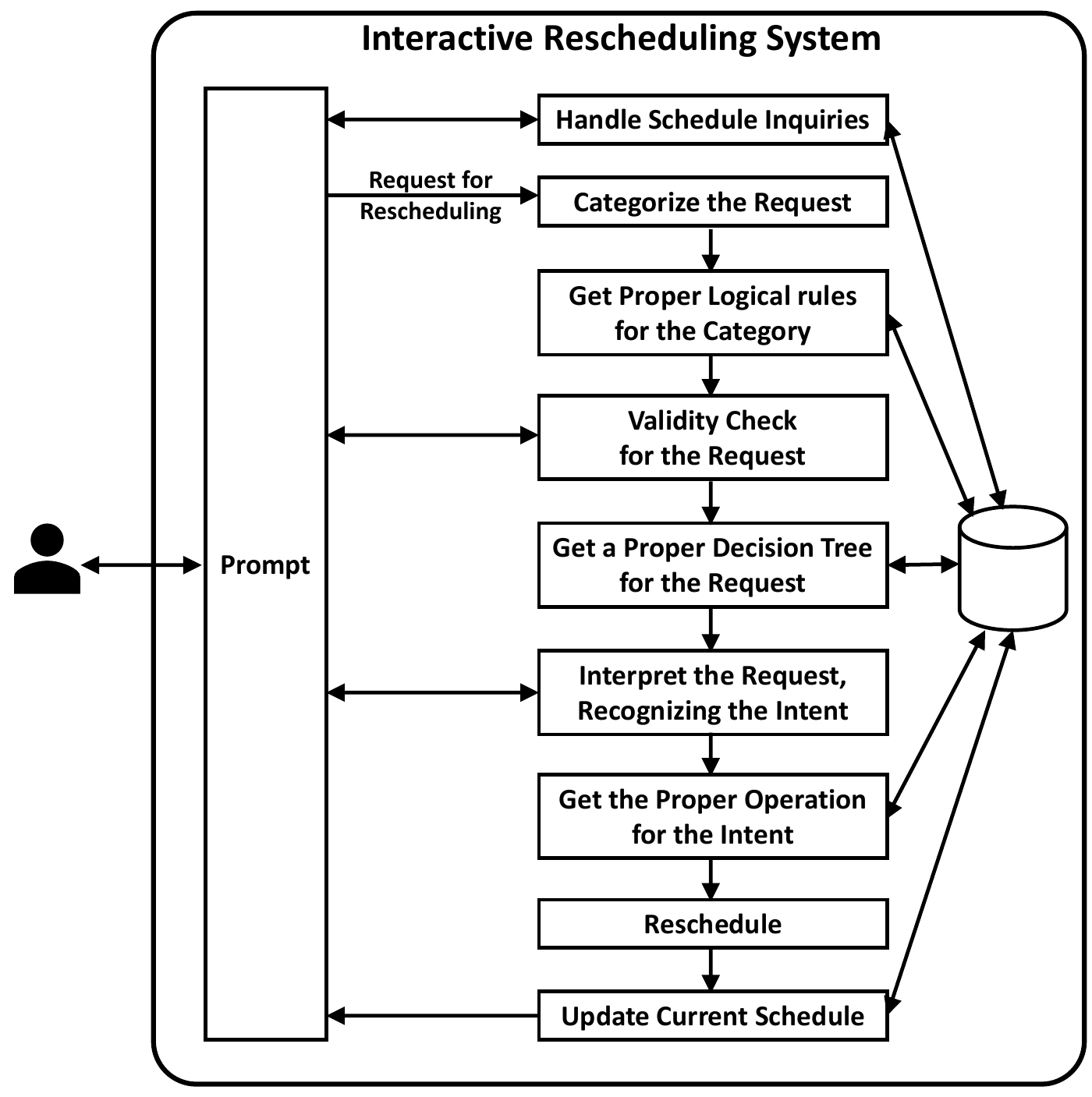}
			\caption{Simple Flow for the highlevel solution}
			\label{fig:sol_overview}
			\vspace*{-0.5cm}
		\end{figure}

%		A simple flow shown in Fig. \eqref{fig:sol_overview} is as follows: %given a set of input pairs as an instance of RCPSP-REQ, we solve the problem repeatedly after reading each pair of the input. When the model returned by the solver is satisfied, we output the result. If we get an unsatisfied result, we print the error. Then we move on to the next input pair. For an iterative approach, as shown in Fig. \eqref{fig:sol_overview}, we repeat the simple flow until the termination condition is met.
%-----
%		Figure \ref{fig:sol_overview} outlines the system's workflow. A user prompt triggers the process. If identified as a rescheduling request, it is categorized and validated using logical rules. An appropriate decision tree then interprets the request, potentially interacting with the user via prompts (updating $\mathcal{H}$) to resolve ambiguity and recognize the final intent ($\vect{b}^{*}$). Based on this intent, the correct rescheduling operation (parameter/constraint adjustments) is selected and executed (solving the RCPSP). The schedule is updated, and the result (new schedule $S$, explanation $E$) is returned. A central database stores rules, trees, and schedule data.
%-----
		Figure \ref{fig:sol_overview} depicts the system's process. A user prompt starts it. Rescheduling requests are categorized and validated using logic. After interpreting the request, a decision tree can interact with the user via prompts to resolve ambiguity and determine the user's intent $\vect{b}^{*}$. This intent identifies and performs the right rescheduling operation (parameter/constraint modifications) to solve the RCPSP. Updated schedule $S$ and explanation $E$ are returned. A central database maintains rules, trees, and schedules.

		\section{Proposed Solution}
		This section describes the method for extracting user intent from potentially ambiguous requests and the subsequent rescheduling process.
		
		\subsection{Intention Extraction}
			The process begins by validating the user request and then interpreting the underlying intention using ASP and an interactive decision tree approach.
		
%		\subsubsection{Checking Validity for the Request}
%		The listing below represents the validity check for a start time change request.
%		\texttt{\scriptsize
%			\begin{align}
%				& request(n, i). \label{eq:validity:postpone_begin}\\
%				valid\_task \leftarrow					& request(N, I), task(N). \\
%				valid\_time \leftarrow					& request(N, I), I \le I2, step(I), \nonumber \\ 
%				& total\_dur(I2).  \\
%				result(valid\_request) \leftarrow 		& valid\_task, valid\_time. \\
%				result(invalid\_task) \leftarrow		& \neg valid\_task, valid\_time. \\
%				result(invalid\_time) \leftarrow		& \neg valid\_time, valid\_task. \label{eq:validity:postpone_end}
%			\end{align}
%		}
%		
%		The listing below represents the validity check for a resource amount change request.
%		\texttt{\scriptsize
%			\begin{align}
%				& request(n, r, a). \label{eq:validity:resource_begin} \\
%				resource(R) \leftarrow				& rsreq(N, R, A). \\
%				limit(R, L) \leftarrow				& rsreq(N, R, A), \nonumber \\
%				& L = \#max\{L1: rsreq(N, R, L1)\}. \\
%				valid\_task \leftarrow				& request(N, R, A), task(N). \\
%				valid\_resource \leftarrow			& request(N, R, A), resource(R). \\
%				valid\_amount \leftarrow			& request(N, R, A1), A1 \le A2, \nonumber \\
%				& limit(R, A2). \\
%				result(valid\_request) \leftarrow	& valid\_task, valid\_resource, \nonumber \\
%				& valid\_amount. \\
%				result(invalid\_task) \leftarrow	& \neg valid\_task. \\
%				result(invalid\_resource) \leftarrow& \neg valid\_resource. \\
%				result(invalid\_amount) \leftarrow	& \neg valid\_amount. \label{eq:validity:resource_end}
%			\end{align}
%		}

		\subsubsection{Checking Validity for the Request}
		ASP rules check if the request parameters (task ID, desired time/resource) are valid within the problem constraints.
		
		The listing below shows the validity check for a start time change request.
		\texttt{\scriptsize
			\begin{align}
				& request(n, i). \label{eq:validity:postpone_begin}\\
				valid\_task \leftarrow					& request(N, I), task(N). \\
				valid\_time \leftarrow					& request(N, I), I \le I2, step(I), \nonumber \\
				& total\_dur(I2).  \\
				result(valid\_request) \leftarrow 		& valid\_task, valid\_time. \\
				result(invalid\_task) \leftarrow		& \neg valid\_task, valid\_time. \\
				result(invalid\_time) \leftarrow		& \neg valid\_time, valid\_task. \label{eq:validity:postpone_end}
			\end{align}
		}
		
		The listing below shows the validity check for a resource amount change request.
		\texttt{\scriptsize
			\begin{align}
				& request(n, r, a). \label{eq:validity:resource_begin} \\
				resource(R) \leftarrow				& rsreq(N, R, A). \\
				limit(R, L) \leftarrow				& rsreq(N, R, A), \nonumber \\
				& L = \#max\{L1: rsreq(N, R, L1)\}. \\
				valid\_task \leftarrow				& request(N, R, A), task(N). \\
				valid\_resource \leftarrow			& request(N, R, A), resource(R). \\
				valid\_amount \leftarrow			& request(N, R, A1), A1 \le A2, \nonumber \\
				& limit(R, A2). \\
				result(valid\_request) \leftarrow	& valid\_task, valid\_resource, \nonumber \\
				& valid\_amount. \\
				result(invalid\_task) \leftarrow	& \neg valid\_task. \\
				result(invalid\_resource) \leftarrow& \neg valid\_resource. \\
				result(invalid\_amount) \leftarrow	& \neg valid\_amount. \label{eq:validity:resource_end}
			\end{align}
		}

		\subsubsection{Interpreting the Intention}
		ASP combined with three-valued logic ($\top, \bot, \mho$ for true, false, unknown) and external atoms (\texttt{\#external sym/2}) models a decision tree for clarifying intent. The system queries the user (\texttt{query/1}) when information is unknown ($\mho$).
		
		The listing below shows the generic decision tree generation logic in ASP for $n$ boolean decision points.
		\texttt{\scriptsize
			\begin{align}
				\#external \quad sym(S, V): \quad				& symbol(S), value(V). 		\label{eq:dec_tree_begin}		\\
				& symbol(1..n).						\\
				& value(``\top";``\bot").			\\
				1\{\top(S);\bot(S);\mho(S)\} \leftarrow	& symbol(S), value(V).				\\
				\mho(S)							\leftarrow	& \neg sym(S, ``\top"), \neg sym(S, ``\bot"), \nonumber \\
				& symbol(S).	 	\\
				contradiction(S) \leftarrow & sym(S, ``\top"), sym(S, ``\bot"), symbol(S). \\
				\mho(S) \leftarrow & contradiction(S), symbol(S). \\
				\leftarrow	& \neg \top(S), sym(S, ``\top"), symbol(S).  \\
				\leftarrow	& \neg \bot(S), sym(S, ``\bot"), symbol(S). \\
				query(S1)							\leftarrow	& O1 = \#min\{O:sym\_order(S, O), \mho(S)\}, \nonumber \\
				& sym\_order(S1, O1), symbol(S1).	\\
				observed(S, V)						\leftarrow	& sym(S, V), symbol(S), value(V).	\\
				assumed(S, ``\neg\bot")				\leftarrow	& observed(S, ``\top"), \neg sym(S, ``\bot"), \nonumber \\
				& symbol(S). \\
				assumed(S, ``\neg\top")				\leftarrow	& observed(S, ``\bot"), \neg sym(S, ``\top"), \nonumber \\
				& symbol(S). \\
				known(S, ``\top")					\leftarrow	& observed(S, ``\top"), assumed(S, ``\neg\bot"), \nonumber \\
				& symbol(S).	\\
				known(S, ``\bot")					\leftarrow	& observed(S, ``\bot"), assumed(S, ``\neg\top"), \nonumber \\
				& symbol(S).	\\
				\leftarrow	& known(S, ``\top"), known(S, ``\bot"), \nonumber \\
				& symbol(S).
			\end{align}
		}
		We assume a default symbol order:
		\texttt{\scriptsize
			\begin{align}
				& sym\_order(i, i). \quad \rhd \textit{$\forall i \in \mathbb{N}^{+}$ s.t. $1 \le i \le n$}
			\end{align}
		}
		Intentions are defined based on the final known states. For $n=2$:
		\texttt{\scriptsize
			\begin{align}
				intent(``op\_00") \leftarrow	& known(2, ``\bot"), known(1, ``\bot"). \\
				intent(``op\_01") \leftarrow	& known(2, ``\bot"), known(1, ``\top"). \\
				intent(``op\_10") \leftarrow	& known(2, ``\top"), known(1, ``\bot"). \\
				intent(``op\_11") \leftarrow	& known(2, ``\top"), known(1, ``\top"). \label{eq:dec_tree_end}
			\end{align}
		}

		Algorithm \ref{alg:interpreter} details the interactive intention interpretation process. It categorizes and validates the request, then iteratively uses the ASP decision tree logic, potentially querying the user via interaction history $\mathcal{H}$, until the intent $\mathbf{b}^*$ (represented by \texttt{known/2} facts) is fully resolved or a timeout occurs. It returns the final intent $\mathbf{b}^*$, request type $\gamma$, and an explanation $E$.
		
		% --- Algorithm 1: InterpretIntention ---
		% (Keep Algorithm 1 exactly as provided in the input file)
		\begin{algorithm}[H]
			\footnotesize
			\caption{InterpretIntention}
			\label{alg:interpreter}
			\SetKwInOut{Input}{Input}
			\SetKwInOut{Output}{Output}
			
			\Input{A baseline schedule $S_0$, a task $t^*$, and an initial request $Q$}
			\Output{An intention $\mathbf{b}^*$, a request type $\gamma$, and explanation $E$}
			Categorize the request type $\gamma$ with given inputs \;
			Check the validity for $Q$ and $\gamma$ with Eq.\eqref{eq:validity:postpone_begin}-\eqref{eq:validity:postpone_end} or Eq.\eqref{eq:validity:resource_begin}-\eqref{eq:validity:resource_end} \;
			\If{\scriptsize{$\neg \textup{\texttt{result(valid\_request)}}$}}{
				\Return{$\left\langle \emptyset, \gamma, \{\scriptsize{\textup{\texttt{result/1}}}\}\right\rangle$}
			}
			$n \leftarrow 2$ \Comment*[r]{\spaceskip.22em Both request $\gamma$ types have the same tree depth.}
			Generate a decision tree $\varDelta$ with $n$ through Eq.\eqref{eq:dec_tree_begin}-\eqref{eq:dec_tree_end} \;
			\texttt{\scriptsize{foundIntention}} $\leftarrow \bot$ \;
			$\mathcal{K}_{base} \leftarrow$ Get current time \;
			$\mathcal{H} \leftarrow \texttt{NIL}$ \Comment*[r]{\scriptsize{Initialize an interactive history}}
			$predicates := \{\}$ \Comment*[r]{\scriptsize{Initialize a predicate dictionary}}
			$\mathbf{b}^* := \left\langle \mho, \mho, \mho, \mho \right\rangle$ \Comment*[r]{\scriptsize{\spaceskip.25em Initialize $\mathbf{b}^*$ to $\mho$s, meaning unknown}}
			\While{$\neg \textup{\texttt{\scriptsize{foundIntention}}}$}
			{
				\uIf{\textup{\texttt{\scriptsize{intent/1}}} is in $predicates$}
				{
					\texttt{\scriptsize{foundIntention}} $\leftarrow \top$ \;
					$E \leftarrow \{\texttt{\scriptsize{intent/1, known/2}}\}$ \;
					Update $\mathbf{b}^*$ with $\gamma$ and $\texttt{\scriptsize{known/2}}$ \;
					continue \;
				}
				\uElseIf{\textup{\texttt{\scriptsize{query/1}}} is in $predicates$}
				{
					\ForEach{$C_{sym} \in predicates[\textup{\texttt{\scriptsize{contradiction/1}}}]$}
					{\spaceskip.28em
						Release external \texttt{\scriptsize{sym/2}} with $C_{sym}$ and $\{``\top", ``\bot"\}$ \;
					}
					$\mathcal{H} \leftarrow \mathcal{H} + \left\langle i, \emptyset \right\rangle$ \Comment*[r]{\scriptsize{\spaceskip.25em The symbol index $i$ from query/1}}
					{\spaceskip.18em
						%Ask the user a query with $H$ and wait to receive for time $\kappa$ \;
						%Based on $H$, Ask the user a question and Wait a response for time $\kappa$. \;
						Based on $\mathcal{H}$, Ask the user and Wait for time $\kappa$. \;
					}
					%					Based on the response, Assign external $\texttt{\scriptsize{sym/2}}$ in $\varDelta$ \;
					Assign external $\texttt{\scriptsize{sym/2}}$ referring to the responded $\mathcal{H}$ \;
				}
				\Else(\Comment*[f]{\spaceskip.20em For the case where the \textit{predicates} is empty}){
					%					{%\spaceskip.11em
						%						
						Assign external $\texttt{\scriptsize{sym/2}}$ referring to the given inputs \;
						%					}
					%					Based on $Q$ and $\gamma$, Assign external $\texttt{\scriptsize{sym/2}}$ in $\varDelta$ \;
				}
				Execute $\varDelta$, Parse the result, and Update the $predicates$ \;
				$\mathcal{K}_{curr} \leftarrow$ Get current time \;
				\If{$\mathcal{K}_{curr} - \mathcal{K}_{base} \ge \textup{\texttt{\scriptsize{TIME\_LIMIT}}}$}
				{
					\texttt{\scriptsize{foundIntention}} $\leftarrow \top$ \;
					$E \leftarrow \{ \texttt{\scriptsize{timeout/0}} \}$ \;
				}
			}
			\Return{$\left\langle \mathbf{b}^*, \gamma, E \right\rangle$}
		\end{algorithm}

\subsection{Pre-processing for RCPSP Solver}
			Algorithm \ref{alg:rcpsp_reschedule_preprocess_inline} prepares the problem for a standard RCPSP solver based on the interpreted intent ($\gamma, b_1, b_2$), desired start time $\tau^*$, and resources $R^*$ for task $t^*$. It modifies the baseline resource requirements ($u^0 \rightarrow u'$) and precedence relations ($S^0 \rightarrow S'$) according to the intent flags. Crucially, it determines the set of tasks $T_f$ that must be fixed at specific start times $\tau_f$. This always includes $t^*$ at $\tau^*$. If $b_1=0$ (local scope), it attempts to fix predecessors of $t^*$ at their baseline times ($S_0$), but only if they don't conflict with $t^*$ or each other and are not required to be flexible due to conflicts propagated from $t^*$. Feasibility checks ensure the requested placement of $t^*$ respects precedence and that the final set of fixed tasks $T_f$ is resource-feasible. The output is a modified problem instance ($\left\langle S', u', T_{f}, \tau_{f} \right\rangle$) suitable for the solver, or an infeasibility status.
			
			% --- Helper Function Descriptions (Concise) ---
			The algorithm uses helper functions:
			\begin{compactitem}
				\item \textfunc{CheckTotalConflicts}: Identifies baseline tasks conflicting resource-wise with $t^*$ at $\tau^*$.
				\item \textfunc{PropagateFlexibility}: Finds all tasks (conflicting ones and their successors) that cannot be fixed at baseline times.
				\item \textfunc{CheckResourceFeasibility}: Verifies if a given set of fixed tasks ($T_f, \tau_f$) is resource-feasible.
				\item \textfunc{BuildSuccessorMap}: Creates a lookup map for task successors based on $S'$.
			\end{compactitem}
			
			% --- Algorithm 2: Pre-process ---
			% (Keep Algorithm 2 exactly as provided in the input file)
			\begin{algorithm}
				\footnotesize
				\caption{Pre-process}
				\label{alg:rcpsp_reschedule_preprocess_inline}
				\SetKwInOut{KwIn}{Input}
				\SetKwInOut{KwOut}{Output}
				\SetKwFunction{CTC}{\scriptsize CheckTotalConflicts}
				\SetKwFunction{PF}{\scriptsize PropagateFlexibility}
				\SetKwFunction{CRF}{\scriptsize CheckResourceFeasibility}
				\SetKwFunction{BSM}{\scriptsize BuildSuccessorMap}
				
				\KwIn{\\
					\begin{itemize}[$\bullet$]
						\item {\spaceskip.30em Request Data: Target task $t^*$, desired time $\tau^*$, desired resources $R^*$, request type $\gamma$, intention $b_1$ and $b_2$ \;}
						\item {\spaceskip.22em Baseline Data: Tasks $T$, Resources $R$, Durations $p$, Precedences $S^0$, Resource Function $u^0$, Capacity $C$, Horizon $H$, Baseline schedule $S_0$ \;}
					\end{itemize}
				}
				\KwOut{\spaceskip.25em
					$\left\langle S', u', T_{f}, \tau_{f} \right\rangle$, or Infeasible status with reason \;
				}
				
				%	\tcc{Step 1: Pre-process Resource Requirements}
				Initialize $u'_{tr} \leftarrow u^0_{tr}$ for all $t, r$ \Comment*[r]{\spaceskip.25em Effective resource reqs.}
				\If(\Comment*[f]{\spaceskip.25em for a resource amount change request}){$\gamma = \text{`r'}$}{
					$u'_{t^*r} \leftarrow R^*_{t^*r}$ for all $r \in R$\;
				}
				
				%	\tcc{Step 2: Pre-process Precedence Relations}
				Initialize $S' \leftarrow S^0$ \Comment*[r]{\spaceskip.25em Effective precedence relations}
				$P^0_{t^*} \leftarrow \{i \mid (i, t^*) \in S^0\}$ \Comment*[r]{\spaceskip.28em Original predecessors of $t^*$}
				$S^0_{t^*} \leftarrow \{j \mid (t^*, j) \in S^0\}$ \Comment*[r]{\spaceskip.28em Original successors of $t^*$}
				\If{$b_2 = 0$}{
					Remove $(i, t^*)$ from $S'$ for all $i \in P^0_{t^*}$\;
					Remove $(t^*, j)$ from $S'$ for all $j \in S^0_{t^*}$\;
					Add $(i, j)$ to $S'$ for all $i \in P^0_{t^*}, j \in S^0_{t^*}$\;
				}
				$Succ' \leftarrow \BSM(S')$ \Comment*[r]{\spaceskip.25em Successor map for $S'$}
				
				%	\tcc{Step 3: Pre-Check 1 - Requested Start Time vs Predecessors (Stricter)}
				$S^{min}_{t^*} \leftarrow \max( \{0\} \cup \{ S_0(i) + p_i \mid i \in P^0_{t^*} \land (i, t^*) \in S' \} )$\;
				\If{$\tau^* < S^{min}_{t^*}$}{
					\Return{Infeasible: $\tau^* <$ earliest predecessor finish time}
				}

				%	\tcc{Step 4a: Initialize Fixed Set with Target Task}
				%			$T_{f} \leftarrow \{t^*\}$ \Comment*[r]{Set of tasks fixed at specific times}
				%			$\tau_{f} \leftarrow \{t^* \mapsto \tau^*\}$ \Comment*[r]{Map of fixed start times}
				$\left\langle T_{f}, \tau_{f} \right\rangle \leftarrow \left\langle \{t^*\}, \{t^* \mapsto \tau^*\} \right\rangle$ \Comment*[r]{\spaceskip.25em Task Set and Start Time Map}
				
				%	\tcc{Step 4b: Find Direct Conflicts within $I_{t^*}$}
				$T_{conf} \leftarrow \CTC(t^*, \tau^*, p, S^0, u', T, R, C)$ \; %Comment*[r]{Conflict Tasks with $t^*$ in $I_{t^*}$}
				
				%	\tcc{Step 4c: Propagate "Must Be Flexible" Status}
				$T_{flex} \leftarrow \PF(S', T_{conf}, T, Succ')$ \; %\Comment*[r]{Tasks required to be Flexible}
				
				%	\tcc{Step 4d: Filter and Fix Predecessors (if $b_2=0$)}
				\If{$b_1 = 0$}{
					%				$C_{fix} \leftarrow P^0_{t^*}$ \Comment*[r]{Candidate predecessors of $t^*$}
					%				$C_{feas} \leftarrow C_{fix} \setminus T_{flex}$ \Comment*[r]{Feasible candidates}
					%				% Sort candidates (optional, for deterministic results)
					%				Sort $C_{feas}$\;
					%\ForEach(\Comment*[f]{For incremental Resource Feasibility}){$i \in C_{feas}$}{
						%				\ForEach{$i \in C_{feas}$}{
							\ForEach(\Comment*[f]{\spaceskip.25em for each feasible candidate}){$i \in P^0_{t^*} \setminus T_{flex}$}{
								%					$\tau^{prop}_i \leftarrow S^0_i$ \Comment*[r]{Proposed baseline start time for $i$}
								%					$T'_{f} \leftarrow T_{f} \cup \{i\}$ \Comment*[r]{Hypothetical Task Set}
								%					$\tau'_{f} \leftarrow \tau_{f} \cup \{i \mapsto \tau^{prop}_i\}$ \Comment*[r]{Hypo. Time Map}
								%			\tcc{Check incremental resource feasibility}
								$\left\langle T'_{f}, \tau'_{f} \right\rangle \leftarrow \left\langle T_{f}\! \cup\! \{i\}, \tau_{f}\! \cup\! \{i \mapsto S_0(i)\} \right\rangle$ \Comment*[r]{Temporarily}
								$f_{R} \leftarrow \CRF(T'_{f}, \tau'_{f}, u', p, C, R, H)$\;
								\If(\Comment*[f]{\spaceskip.25em Task ${i}$ is resource-wise feasiable}){$f_{R}$}{
									%						$T_{f} \leftarrow T'_{f}$\; % Update accepted fixed tasks
									%						$\tau_{f} \leftarrow \tau'_{f}$\; % Update accepted fixed times
									$\left\langle T_{f}, \tau_{f} \right\rangle \leftarrow \left\langle T'_{f}, \tau'_{f} \right\rangle$ \Comment*[r]{\spaceskip.25em Add task $i$ Permanently}
									%				\tcc{Commit to fixing task i}
									%			}\Else{
									%				\tcc{Cannot fix i due to resource conflict with other fixed tasks}
								}
							}
						}
						
						%	\tcc{Step 5: Final Resource Feasibility Check for the Entire Fixed Set}
						$f_{final} \leftarrow \CRF(T_{f}, \tau_{f}, u', p, C, R, H)$ \;
						\If{$\neg f_{final}$}{
							\Return{Infeasible: Final fixed set $T_{f}$ violates resources}
						}
						
						%	\tcc{Return successful results}
						\Return{$\left\langle S', u', T_{f}, \tau_{f} \right\rangle$}
						
					\end{algorithm}

			\subsection{Rescheduling Execution}
			The overall rescheduling process is managed by a dispatcher mechanism. After \textproc{InterpretIntention} (Alg. \ref{alg:interpreter}) determines the request type $\gamma$ and intent flags $(b_1, b_2)$, the dispatcher selects the appropriate logic flow. This involves calling the \textproc{Pre-process} function (Alg. \ref{alg:rcpsp_reschedule_preprocess_inline}) with the specific intent flags. If preprocessing is successful and returns a feasible modified instance ($\langle S', u', T_f, \tau_f \rangle$), the dispatcher formulates this as an RCPSP problem with fixed tasks and passes it to a standard RCPSP solver. If preprocessing fails or the solver returns infeasible, the dispatcher reports the failure and the reason derived from the preprocessing or solving stage. Otherwise, the final schedule returned by the solver is the output. This structure allows different combinations of user intentions to be handled by the same core preprocessing and solving steps, varying only the parameters passed to \textproc{Pre-process}.

            \subsection{Formal Verification of the Interactive Mechanism}
			\label{subsec:formal_proofs}
			
			In order to build trust via human-automation interaction, we show the formal guarantees on the termination of the intent extraction process and its correctness. We here define the status of the user's intent in the domain $\{\top, \bot, \mho\}$ at step $k$ as vector $\mathbf{v}_k$.
			
			\begin{theorem}[Convergence]
				Let $\mathcal{S}$ denote a set of $n$ decision symbols required to completely specify the request. Assuming that the user always answers inquiries from the system in the given time, the InterpretIntention algorithm terminates in at most $n$ steps of interaction.
			\end{theorem}
			
			\begin{proof}
				Let $U_k = \{s \in \mathcal{S} \mid \text{value}(s) = \mho\}$ denote a set of undecided symbols at iteration $k$. Since initially $U_0 = \mathcal{S}$, $|U_0| = n$. This is executed iteratively. In the While loop in Alg.~\ref{alg:interpreter}, the system identifies the symbol $s^* \in U_k$ and generates an inquiry. When a response $a \in \{\top, \bot\}$ (or the timeout) is returned, the ASP solver derives the predicate $\text{known}(s^*, a)$. Thus, the rule for the default negation of $\mho$ does not hold any longer as follows:
				\begin{equation}
					U_{k+1} = U_k \setminus \{s^*\}
				\end{equation}
				
				This implies $|U_{k+1}| = |U_k| - 1$.
				Then, the cardinality of the set becomes strictly decreasing until it reaches 0, which is the lower bound.
				When $|U_k| = 0$, the set of all symbols turns to a known state.
				The ASP rule (Eq.~\ref{eq:dec_tree_end}) for \texttt{intent} requires the conjunction of every $s \in \mathcal{S}$ to be a known state.
				When $|U_k| = 0$, this condition holds, and then it derives \texttt{foundIntention}, terminating the loop.
			\end{proof}

			\begin{theorem}[Correctness]
				The intent code $\mathbf{b}^*$ derived from the system is exactly matched with the conjunction of the user's sequence of answers.
			\end{theorem}
			
			\begin{proof}
				The decision tree $\varDelta$ is to implement the boolean function $f: \{0,1\}^n \to \mathcal{I}$, and here $\mathcal{I}$ is a set of intent labels. ASP encoding enforces that the observed history $\mathcal{H}$ is in a bijection relation between internal states. For an arbitrary input vector $\mathbf{a} = (a_1, \dots, a_n)$, the interactive loop guarantees  the assertion of a set of facts $F = \{\text{known}(i, a_i) \mid 1 \le i \le n\}$. The intent rules have the following form:
				\begin{equation}
					\text{intent}(L) \leftarrow \bigwedge_{i=1}^n \text{known}(i, v_i)
				\end{equation}
				
				Since stable model semantics guarantees that the consequence is derived if and only if the bodies are satisfied, a unique intent label $L$ that is corresponding to $\mathbf{a}$ is derived. The constraint \texttt{contradiction/1} guarantees logical consistency by restricting a conflict state (for example, setting to $\top$ and $\bot$ simultaneously) from occurring.
			\end{proof}

				\section{Experiments and Discussion}
				
%				 \begin{figure}
%					\centering
%					\begin{subfigure}[b]{0.8\linewidth}
%						\includegraphics[width=\linewidth]{images/fig_2_tree_view_cropped}
%						\caption{\spaceskip.18em Precedence Relations with Two Dummies: 0 and 11.}\label{fig:toy:trees}
%						\vspace{0.8\baselineskip}
%					\end{subfigure}
%					
%					\begin{subfigure}[b]{.493\linewidth}
%						\includegraphics[width=\linewidth]{images/fig_2_original_sol2_cropped}
%						\caption{Resource Assignment for the Baseline Solution}\label{fig:toy:baseline}
%					\end{subfigure}
%					\begin{subfigure}[b]{.493\linewidth}
%						\includegraphics[width=\linewidth]{images/fig_2_rescheduled_sol2_cropped}
%						\caption{Resource Assignment for the Rescheduled Solution}\label{fig:toy:rescheduled}
%					\end{subfigure}
%					\vspace{0.4\baselineskip}
%					\caption{A Comparison between the Baseline Schedule and the Rescheduled One for the Resource Change Request}
%					\label{fig:toy_example_result}
%				\end{figure}	

	\subsection{Case Study: Rescheduling Task 4}
    
We demonstrate rescheduling using the instance from Table~\ref{tab:rcpsp_example_condensed_reduced} (baseline schedule shown in \figtoybaseline{}). A user requests changing Task 4's start time to $\tau^*=4$ (from baseline 5) and reducing Resource 1 usage to $R^*_{4,1}=1$ (from baseline 2). The interaction shown in Listing \ref{lit:result:intention} clarifies the intent as $\gamma = \text{'r'}$ (resource-focused), $b_1 = 1$ (allow global optimization), and $b_2 = 1$ (keep precedence links).
	
\label{subsec:case_study}
				\vspace{-0.2cm}
				\begin{lstlisting}[basicstyle=\ttfamily\scriptsize, label=lit:result:intention, caption={Interactive Inquiries to Reveal the User's Intent.}, frame=single]
Result:request(4,1,1) result(valid_request)

===== interaction #00 =====
unknown: [unknown(1), unknown(2)]
query: [query(1)]

===== interaction #01 =====
Do you want the new schedule to optimize by affecting 
other tasks' start times? Yes for 1; no for 0.
?[1, 0]:Yes
Invalid answer: Yes

Do you want the new schedule to optimize by affecting 
other tasks' start times? Yes for 1; no for 0.
?[1, 0]:2
Invalid answer: 2

Do you want the new schedule to optimize by affecting 
other tasks' start times? Yes for 1; no for 0.
?[1, 0]:1
sym: [sym(1,"true")]
unknown: [unknown(2)]
query: [query(2)]

===== interaction #02 =====
Do you want the task to depend on others while mainta-
ining precedence? Yes for 1; no for 0.
?[1, 0]:1
sym: [sym(1,"true"), sym(2,"true")]
intent: [intent(op_11)]
\end{lstlisting}

%-----
%	The pre-processing step updates $u'_{4,1}=1$, fixes only Task 4 ($T_{\textup{f}}=\{4\}$) at $\tau_4=4$, and retains its precedence relations ($S'$ unchanged locally). After successful feasibility checks, the modified RCPSP instance is solved. The resulting schedule (\figtoyrescheduled{}) shows Task 4 starting at $t=4$ using 1 unit of Resource 1, leading to a new makespan of 21. This illustrates the system's ability to generate feasible, intent-aligned schedules.
%-----
	After updating $u'_{4,1}=1$ in the pre-processing step, Task 4 ($T_{\textup{f}}=\{4\}$) is fixed at $\tau_4=4$, and precedence relations ($S'$) are maintained locally. After successful feasibility checks, it solves the updated RCPSP instance. As seen in \figtoyrescheduled{}, Task 4 begins at $t=4$ and uses 1 unit of Resource 1, resulting in a new makespan of 21. This shows that the system is capable to generating feasible, intent-aligned schedules.

\subsection{Discussion}
\label{subsec:discussion}

Our technique enables intention-aware rescheduling but suggests improvements. While visualizations like \figtoybaseline{} and \figtoyrescheduled{} aid user communication, the binary intent flags ($b_1, b_2$) may simplify complicated user priorities, causing usability issues.

In order to handle an infeasible plan, more is required than simply reporting its failure. Proposing a feasible alternative or pinpointing the exact conflict point can be constructive and useful to the user even though the computational cost would be high.

In addition, the current offline approach should be extended to dynamic settings. Adapting the current framework for online rescheduling techniques such as rolling horizon planning and localized repair is necessary. However, this extension requires a balance between the intent satisfaction, feasibility, and responsiveness.
				
				\section{Conclusions}
\label{sec:conclusions_condensed}

This research proposed an intention-aware rescheduling methodology in order to handle ambiguous human requests in UAM vertiport operations in RCPSP settings. For the proposed solution, we showed the integration of ASP-driven intent interpretation through three-valued logic and MILP-driven optimization. ASP provides transparency, builds trust, and enables an explainable rescheduling system.

Computational scalability, deeper intent modeling, and online, dynamic scheduling for real-time UAM should be addressed in the future.  Practicality and user approval must be confirmed by thorough usability research.

\section*{ACKNOWLEDGMENT}

This research was supported by the Regional Innovation System \& Education (RISE) program through the Jeollanamdo RISE center, funded by the Ministry of Education (MOE) and the Jeollanamdo, Republic of Korea. (2025-RISE-14-008)

\bibliographystyle{IEEEtran}
\bibliography{ref}

\end{document}